# Understanding Grasp Synergies during Reach-to-grasp using an Instrumented Data Glove

Subhash Pratap, *Student Member, IEEE*, Yoshiyuki Hatta, *Member, IEEE*, Kazuaki Ito, *Senior Member, IEEE*, Shyamanta M. Hazarika, *Senior Member, IEEE*

*Abstract*—Data gloves play a crucial role in study of human grasping, and could provide insights into grasp synergies. Grasp synergies lead to identification of underlying patterns to develop control strategies for hand exoskeletons. This paper presents the design and implementation of a data glove that has been enhanced with instrumentation and fabricated using 3D printing technology. The glove utilizes flexible sensors for the fingers and force sensors integrated into the glove at the fingertips to accurately capture grasp postures and forces. Understanding the kinematics and dynamics of human grasp including reach-to-grasp is undertaken. A comprehensive study involving 10 healthy subjects was conducted. Grasp synergy analysis is carried out to identify underlying patterns for robotic grasping. The t-SNE visualization showcased clusters of grasp postures and forces, unveiling similarities and patterns among different GTs. These findings could serve as a comprehensive guide in design and control of tendon-driven soft hand exoskeletons for rehabilitation applications, enabling the replication of natural hand movements and grasp forces.

## I. INTRODUCTION

The human hand is a crucial component in daily life, enabling individuals to interact with and manipulate their environment through grasping objects.

It is a complex end effector allowing for over 20 degrees of freedom. There are 27 bones in the human hand, including 8 small bones called carpals, 5 metacarpal bones, and 14 phalanges. These bones are connected by joints that allow for movement and flexibility. The muscles of the hand are arranged in groups, with each group controlling a specific movement or action. The tendons connect the muscles to the bones and allow for transfer of force to the hand and fingers. The human hands and fingers posses a notable degree of dexterity and are proficient at executing a diverse range of fine motor actions, enabling us to engage in various daily living activities (DLAs) [1].

However, only a reduced number of manipulation postures are consistently used in our day-to-day lives. A human performs on average 4000 to 5000 grip changes during a typical 8-hour work day [2]. Grasp-hold-release tasks are well recognised as being one of the most prevalent operations involved in various DLAs by hand [3]. Hand grasp patterns require complex coordination. The term "synergy," derived from the Greek word "synergia," signifies the concept of elements working together towards a common objective [4].

Grasp synergy can be defined as a fundamental spatial hand configuration that is commonly observed across a wide range of tasks. The complex biomechanical and neurological structure of hand articulation gives rise to complex questions surrounding the control mechanisms that govern the synchronization of finger movements and the grip forces necessary for manipulating a wide range of objects in DLAs. This intricacy spans a wide range of tasks, including the grasping of objects using several digits and the fine motor control of individual fingers. Out of the five fingers and more than 20 joints in a human hand, it's worth noting that, theoretically, only three fingers and nine joints are enough for the purpose of grasping and manipulating objects [5]. The issue of this surplus, often referred to as redundancy, has been a fundamental focus in the field of motor control research from its inception [6]. By adjusting the flexibility of the human hand, it becomes possible to apply multiple combinations of force synergies while maintaining the identical postural synergy representation. This deliberate manipulation allows for the intentional control of grasp stability and the total force exerted on an object by an individual.

The dexterity of the human hand is frequently compromised by movement disorders like Parkinson's disease and paralysis caused by conditions such as stroke and spinal cord injury. Assessing hand function in these individuals is a crucial aspect of clinical evaluation, which can aid in diagnosis, treatment, and rehabilitation. Understanding the natural way of human grasp, insight into the kinematic indications and constraints associated with each grasp, and knowledge of frequently used patterns hold significance across diverse domains, including medicine, rehabilitation, psychology, and product design, among numerous others.

This paper describes the design and development of a 3D-printed instrumented data glove. The data glove records grasp forces and postures. We present a novel approach to represent the grasp synergies within the postural and force domains. These postural and force synergies serve to depict

Subhash Pratap is Joint Degree Research Scholar associated with the Biomimetic Robotics and Artificial Intelligence Laboratory (BRAIL), Department of Mechanical Engineering, Indian Institute of Technology Guwahati (INDIA) and Motion Laboratory, Department of Intelligent Mechanical Engineering, Gifu University (JAPAN). e-mail: subhash.iitg18@gmail.com

Yoshiyuki Hatta and Kazuaki ITO are with the Motion Laboratory, Department of Intelligent Mechanical Engineering, Gifu University (JAPAN). e-mail: hatta.yoshiyuki.b3@f.gifu-u.ac.jp; ito.kazuaki.x5@f.gifu-u.ac.jp

Shyamanta M. Hazarika is with the Biomimetic Robotics and Artificial Intelligence Laboratory (BRAIL), Department of Mechanical Engineering, Indian Institute of Technology Guwahati, India, Pin: 781039 e-mail: s.m.hazarika@iitg.ac.in

grasping in a reduced-dimensional space and draw inspiration from the inherent correlations between finger grasp postures and the forces they exert. To assess the effectiveness of this novel representation, we executed a human grasping experiment involving ten participants. This experiment involves the manipulation of objects typically encountered in daily living activities, with varying objects of different sizes and masses, and encompasses the use of eight distinct grasp types (GTs). Using a custom built data glove, we capture contact forces at the fingertips of the hand together with the fingers bending angle values.

The main contributions of this article are as follows:

a. *Custom made Data Glove:* Fabrication of a 3D printed data glove embedded with a combination of thin and flexible sensors capable of measuring grasp postures and fingertip forces; accommodating different hand sizes.
b. *Insights into Grasp Force and Grasp Posture Relationship:* Extensive experiments provided insight into the relationship between grasp force and grasp posture in human hand grasping of objects used in DLAs, including the influence of object mass on grasp force and finger cooperation characteristics.
c. *Exploration of Grasp Synergies:* A detailed analysis of postural and force synergies is done for 8 different grasp types involved during DLAs in terms of force and postures and then the grasp types are grouped using clustering techniques for better understanding of the underlying patterns by unveiling similarities among different GTs.

The remainder of this article is organized as follows. In Section II, we review the status of grasp synergies, with special attention to reach-to-grasp and how data gloves are used to understand the same. The detailed description of the modeling and analysis for the data glove is provided in Section III, while Section IV discuss the design and prototype of the data glove. Section V reports extensive experiments and analysis. Results and Discussion is presented in Section VI. We gave a brief conclusion in Section VII.

## II. Literature Review

The way in which we interact with our environment through our hands has been the subject of many studies and classifications. Schlesinger was the first to identify six common grasp patterns, including cylindrical, tip, hook, palmar, spherical, and lateral grasps [7]. The posture of the hand being influenced by the shape and size of the object being manipulated. Napier later added that the nature of the task also plays a role, with some grasps providing stability and intensity (power grasps), while others provide dexterity through the thumb and fingertips (precision grasps)[8]. Cutkosky differentiated power and precision grasps and considered shape and size, as well as non-prehensile postures[9]. Bullock created a more extensive taxonomy which included subtle within-hand movements[10]. Other researchers have focused on the geometry of grasps, such as Lyons who introduced the concepts of encompass, lateral, and precision grasps, and the concept of a virtual finger [11]. A grasp taxonomy with 33 digfferent grasps have been proposed[12]. However, grasp usage depends on the context and not all grasps are used equally, with the index finger and thumb being the most used digits, while the frequency of use of the other fingers decreases from the middle finger to the little finger. Instrumented data gloves play a pivotal role in accurately measuring human grasp by detecting hand motion, finger bending, and force feedback. Their significance spans a wide range of applications, from enhancing human-machine interaction to facilitating biomedical research [13].

### A. Assessing Human Grasp Functionality using Instrumented Data Glove

Assessment of human grasp functionality requires two key steps: a) classification of hand movement patterns (grasp postures), b) measurement of grasp forces at fingertips. Several techniques have been proposed for capturing the grasp postures and grasp forces. In general, all of these approaches for collection of the human grasp posture data can be grouped under two approaches: vision-based approaches and wearable device based (data gloves) [13]. Vision based systems track a user's hand positions using optical markers. Occlusion continues to be a major drawback of optical-based devices[14]. Marker placement mistakes due to skin deformation and marker movements, restricted measurement space, specialized cameras, and heavy markers or suits are some of the challenges of vision based approaches for measuring grasp postures [15] Their identification accuracy ranges from 69% to 98% [16]. Compared with the vision-based devices, the wearable ones rely only on intrinsic sensor readings, which offers better robustness to the environment and has achieved about 90% accuracy on an average[17] [18].

Data gloves are specialized types of gloves equipped with inbuilt sensors that are designed to detect and capture hand motions, joint angles, finger movements like flexion-extension, and even haptic feedback information across the finger pads to provide a detailed representation of grasping and manipulation [19].These gloves employ electronic sensors to detect hand gestures, and at times the amount of force applied to serve as a human- machine interface. Data gloves offer high precision to accurately capture hand motion, rendering them a highly important tool for tasks that require fine-grained control and dexterity. Sensor-based systems are therefore more feasible to be used than optical-based ones [20].Due to their lightweight, compactness, affordability, and durability, flex sensors and inertial measurement units (IMUs) are the two primary types of sensors used in measuring the grasp postures [16][21][22]. G. Saggio et al. [23] found that gloves with resistive flex sensors (RFSs) and inertial measurement units (IMUs) are both suitable for dynamic interactions and yield comparable results to static or quasi-static assessments. Despite being well developed and applicable to a variety of applications, the methods for evaluating hand kinematics cannot account for complete information of human grasp. Several studies have used force sensing along with motion sensors to detect the fingertip force as well as the kinematics of the hand [24][25][26]. Hsiao et al. [24] developed a data glove integrating nine-axis IMUs and force sensing resistors (FSRs) for the evaluation of hand function, however, this study didn't thoroughly validate the reliability of joint angles and fingertip force, especially since the FSRs used were non-linear.To measure finger-object interactions, Kortier et al.[25] proposed a device that used force sensors in the data glove. However, the bulky force sensor may not be appropriate for fine-grained rehabilitation activities. Liu et al. [26] introduced a glove-based system that combined IMUs with force sensors to study hand-object manipulation. Eventhough lighter and more comprehensive force sensor than previous attempts were used they failed to verify the accuracy of hand force and kinematics measurements. Zheng et at. [27] proposed an e-glove for analysing a damage-free manipulation for bionic manipulators. Inspite of the possibility of simultaneously capturing hand kinematics and fingertip force being available, most data gloves lached any significant development in this direction, highlighting the need for a system to address these issues. These limitations include the inadequacy of most designs to accommodate various hand sizes, their bulkiness and discomfort when worn, sensor selection lacking proper comparison, and the challenge of integrating established sensorized gloves for hand kinematics with fingertip force measurement, a critical factor in assessing human grasp function. To overcome these challenges, this study introduces a novel 3D printed flexible data glove instrumented with resistive flex sensors and capacitive force sensors, enabling the collection of syncronized kinematic data and fingertip force measurements. The proposed data glove is light in weight, comfortable to wear, biomechanical aligned with the human hand, and adjustable to fit a range of hand sizes.

*B. Grasp Analysis: Postural and Force Synergy*

The investigation of grasp synergies concerning hand postures and grasp forces constitutes a highly dynamic area of research. The data collected through the glove is typically processed using signal processing techniques to extract pertinent features. Dimensionality reduction methods are applied to identify grasp synergies [13].

Numerous studies have focused on hand health assessment and complementary rehabilitation using data gloves [19], [28], [29] . In these application areas, gesture recognition with gloves has also become mainstream and is used for accuracy validation [30], [31], [32] . Starke et al. [33] proposed the concept of force synergies as an approach to express grasp synergies inside the force space. This framework allows for the concise depiction of grasp forces using a reduced number of dimensions. Olikkal et al. [34] added that combining kinematic and muscle synergies through data fusion improves movement reconstruction, suggesting their potential for enhancing the control of prosthetics and exoskeletons. In [35], grasp postural synergies were extracted from a large publicly available database of kinematic hand grasps using cyberglove.

Liu et al. [36] explored the cooperative elements of hand grasping using postural synergy extraction, hand action simplification, and data dimension reduction.

In most of the above-mentioned works, the authors have focused on a single type of synergy *i.e.* either postural, muscle or force synergy.

The instrumented data glove designed and developed in this paper with minimal number of sensors measure grasp postures and fingertip forces, deriving postural and force synergies. This is through a human grasping study using the instrumented data glove and dimensionality reduction techniques. Relationship between grasp force and bending angles, the influence of object mass on grasp force, and finger cooperation characteristics is also studied. Grasp synergy analysis undertaken in this paper holds promise for developing control algorithms that adapt to different grasp tasks and user intentions, aligning with the evolving field of assistive technology.

III. MODELING AND DESIGN OF THE DATA GLOVE

Data glove-based systems for grasp synergy extraction involve a basic architecture (refer Figure 1) consisting of a data glove equipped with sensors to capture hand kinematics. These sensors integrate flex sensors, strain gauges, and force sensors etc to measure hand movements and tactile information accurately. The acquired data is processed using signal processing techniques to extract relevant features, and dimensionality reduction techniques are applied to identify grasp synergies.

*A. Mathematical Representation of Human Hand Finger*

The development of the Data Glove incorporates a mathematical model structure that entails the measurement of the Metacarpophalangeal (MCP) joint, Proximal Interphalangeal (PIP) joints and Distal Interphalangeal (DIP) joint. The hand has one-axis constraints, meaning a single sensor is sufficient to measure the movement of a PIP, DIP, as well as the MCP joint of the same digit. This approach allows for a reduction in the number of sensors required while still having the ability to measure 10 degrees of freedom (DOF) of the hand. Despite the hand having multiple DOF, finger movements are highly constrained and cannot perform arbitrary gestures due to motion constraints. For example, fingers cannot bend too far backward. The assessment of the range of motion (ROM) of phalanges is a crucial aspect of clinical practice. Healthcare practitioners frequently employ universal goniometers, inclinometers, or electro-goniometers to evaluate the declination angles of finger joints for assessing the range of joint movement [37]. Figure 3 illustrates the configuration of each joint and the angles of intrest for the finger joints.

According to Metcalf's hypothesis introduced in their work [38] When the fixed base position of the metacarpophalangeal (MCP) joint is established, there is a unique combination of MCP, PIP, and DIP joint angles required to position the fingertip in a specific location. The Carpometacarpal Joints (CMC) of the thumb facilitates a rotation of 10 to 15°, abduction and adduction of 40 to 80°, and flexion and extension of 50 to 80°. The broad range of motion afforded by the thumb CMC is a significant contributor to the hand's functionality in humans. [39]. The four fingers have IP joints that allow for a range of motion of 110° at the PIP joint and 90° at the DIP joint. The thumb also has an IP joint that allows for a similar range of motion. [39]. The mathematical description

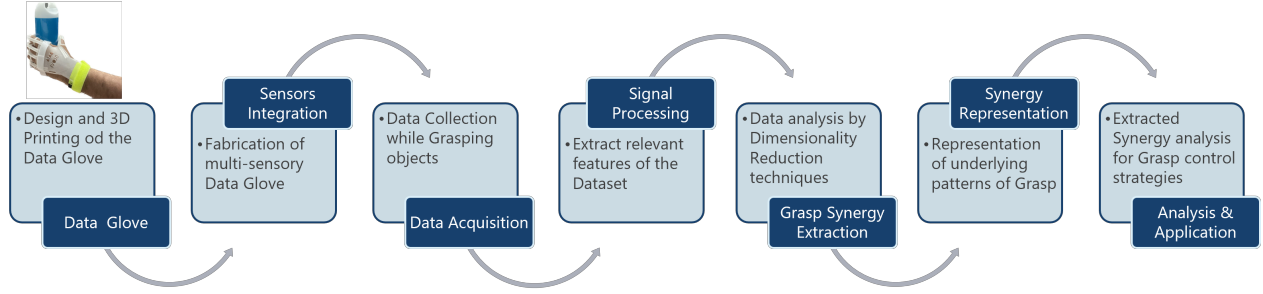

Fig. 1: The fundamental structure of data glove-based systems for kinematic and dynamic synergy study of Grasp

of the normal behaviour of human hand finger movements was described by [40], (refer figure 3),

$$\theta_{DIP} \approx \frac{2}{3}\theta_{MCP}, Fingers \in \{I, M, R, P\}$$

$$\theta_{PIP} \approx \begin{cases} \frac{3}{4}\theta_{MCP}, & Fingers \in \{I, M, R, P\} \\ \frac{1}{2}\theta_{MCP}, & Finger \in \{T\} \end{cases}$$

where T, I, M, R, and P refer to the Thumb, Index, Middle, Ring and Pinky fingers respectively.

Although [41] confirmed this hypothesis through experiments, the study was conducted solely on healthy individuals. It is uncertain whether all the joints and muscles in an injured hand will adhere to the usual constraints and movements. A precise and standardized tool is essential in accessing the active range of motion of the hand in any progressive assessment scenario within the field of hand therapy practice. This is part of the research objective.

### B. Design of the Data Glove

The novelty of the data glove can be attributed to several major aspects that have been carefully considered in its design. The desiderata of the glove design is listed below. Overall, the careful consideration of these major aspects has resulted in a novel and effective data glove design.

1. One important aspect is the generic product design, which would enable the glove to be used across a range of applications and with various hand sizes.
2. Selection of flexible material selection allow the glove to conform to the shape of the hand and provide a comfortable fit for the wearer.
3. The sensors selection is also a critical aspect. Specifically selected flexible sensors that measure both finger flexion/extension and fingertip force have been incorporated into the design. These sensors are placed in locations that accurately capture the intended movements and forces during grasping.
4. Wearability is a crucial aspect of the glove design. The glove has been designed to be lightweight and unobtrusive, with minimal interference to the wearer's natural hand movements. This ensures that the data collected during experiments accurately reflects real-world grasping behavior.

In the instrumented glove system described in this work, the purpose is not to physically assist the user, but rather to measure the grasp characteristics in terms of postures and forces at each finger while grasping objects of daily living activities (DLAs). Deterministic factors of grasp force quality can be analyzed to identify a secure grasp [42]. Detecting the stability of a grip is a challenging aspect in robot-assisted grasping tasks. Poor contact forces can lead to slippage of the grasped object during a manipulation operation, which further complicates the problem. The design involves a thin layer of flexible plastic material (Thermoplastic polyurethane-TPU) containing sensors embedded on top of the finger, which can extend as the finger flexes, similar to a tendon. The dorsal side of the finger consists of grooved T-shaped slots which provide a passage for assembling the flex sensors at the top. Starting from the DIP joint to the tip of the finger, a cap-like structure is designed. The fingers are inserted in these caps through the ring-like structures which are designed at the phalanges. It gives a proper alignment of the data glove across the fingers while the grasping experiment.

### IV. PROTOTYPE OF THE DATA GLOVE

#### A. Fabrication of the Data Glove

3D printing has been used to produce the glove, which allows for precise control over the sensor placement and ensures consistency across multiple copies of the glove. This method also enables the fabrication of complex geometries that would be difficult to achieve using traditional manufacturing techniques. Raise3D Premium TPU-95A (Thermoplastic polyurethane) is a flexible and elastic 3D printing filament. The TPU material's rubber-like properties which include its elasticity, resilience, and durability, make it well-suited for applications that require flexibility in printing and a surface with a soft tactile sensation. The flexible TPU material has a tendency to stick to itself during the printing process, especially when printing structures with overhangs or bridges.

This can lead to distorted or non-functional printing with an irregular surface finish. To overcome this problem, support structures are incorporated during the printing process. However, when these support structures are created using the same TPU material, they also adhere to the print, making the removal of the support structure challenging and risking damage to the final product. To avoid this, a different material like PLA (Polylactide) is employed for printing the support structures. PLA is a rigid and brittle material, which makes its removal easy from the final product without causing any damage. Furthermore, PLA has a higher melting point compared to TPU, which makes it withstand the heat of the printing process without sticking with the TPU material. By opting for PLA as the material for support structures, successful TPU printing can be achieved without any adhesion problems, and the support structures can be effortlessly removed post-printing to achieve the intended final product.

## B. Glove Hardware

The glove hardware include a 3D printed glove embedded with Finger Tactile Pressure Sensors (FingerTPS, Pressure Profile Systems, Los Angeles, CA, USA) and Flex (Spectra Symbol, Salt Lake City, UT, USA) sensors across each digit. The force sensors are attached in the form of caps in the fingers with their sensing element at the tip of the finger The force sensor is equipped with synchronised video recording capabilities, enabling the correlation of tactile data with visual representations of the original use case. The Chameleon Software developed by PPS enables the real-time capture and display of accurate force data and video images. The flex sensors are thin and aligned only on the dorsal side of the fingers, which minimises perceived changes in hand dexterity. These sensors are fitted in the grooved slots of the 3D-printed design of the glove. The sensors are connected to an instrumentation board programmed with Arduino. The complete setup of the 3D-printed fabricated data glove is shown in Figure 2.

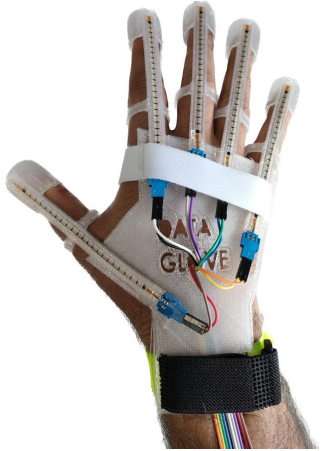

Fig. 2: The 3D printed Data Glove setup

The system consists of Force sensors which is connected with its Chameleon software interface and flex sensors are connected to the Arduino. These sensors are placed on the glove as shown in Figure 2. The flex sensors are placed at the dorsal side of the fingers and are kept into a fixed thin tube for its pallalel movements along the fingers while grasping. While grasping, the flex sensors adjusts itself with the bending of fingers. The FingerTPS force sensors are worn as a cap over the gloves and it measures the grasp forces from the finger tips.

*Finger Bending Angle Measurement:* The multisensory glove is meticulously crafted to incorporate five flex sensors, carefully positioned and oriented along the dorsal surface of the fingers. These sensors actively capture and transmit real-time data regarding the extent of finger movement or Range of Motion (ROM) during grasping. The data glove is meticulously designed to offer a secure and form-fitting experience for each finger. This design ensures that the flexing of individual digits is accurately replicated by the glove's shape. Additionally, the glove provides the convenience of easy hand insertion, reducing the need for intricate hand movements during the donning process.

Table I
SPECIFICATION OF FLEX SENSOR.

| Parameter | Value |
|---|---|
| Length | 4.5 inch (11.43 cm) |
| Flat Resistance | 10K Ohms |
| Resistance Tolerance | ±30% |
| Length | 12 cm |
| Thickness | 1 mm |
| Bend Resistance Range | 60K to 110K Ohms |
| Power Rating | 0.50 Watts continuous, 1 Watt Peak |
| Life Cycle | > 1 million |

The specification of flex sensor used in the data glove (4.5 inch) is shown in Table I. It is imperative to acknowledge that for a human hand that is functioning within normal parameters, the *MCP* joints exhibit autonomous operation from both the *DIP* and the *PIP* joints. In addition, it is commonly seen that the *DIP* and *PIP* joints exhibit concurrent flexion. As a result, the use of the Flex Sensor in this particular scenario may not yield an assessment of the specific angle at which a finger is flexing, as it fails to consider the individual articulations of each joint. In contrast, it captures the comprehensive flexion encompassing the angles of the *MCP*, *DIP*, and *PIP* joints and interprets the overall combination of the three joint angles. The thumb, in a similar manner, integrates the angles of the *IP* and *MCP* joints. Figure 3 presents a schematic depiction of the Flex Sensor, showcasing its initial position at 0° (equivalent to an angle of 180°) and its completely flexed condition at 0°.

The electronic hardware design utilised the Arduino UNO open-source microcontroller as its foundation, enabling the collection of flex sensor angles, encompassing the combined angles of *MCP*, *DIP*, and *PIP* flexion/extension. A configuration consisting of five resistive flex sensors is linked to a conventional 5 V voltage supply. As indicated in Figure 3 (A), the resistance of a flex sensor is measured to be 25 Kohms when it is in its flat or resting orientation. This corresponds to the state where the *MCP*, *DIP*, and *PIP* joints are at rest, as shown in Figure 3 (B). The flex sensor's resistance increases when it undergoes flexion, that is when the finger is in a state of bending. In the state of complete flexion of the sensor, similar to the fully flexed position of a finger as depicted in Figure 3 (A), the resistance of the flex sensor increases to 100 kiloohms, indicating a pinch bend angle of 0 degrees. As a result, the flex sensor demonstrates a high level of proficiency in accurately detecting and recording the full spectrum of finger movements, including both flexion and extension movements.

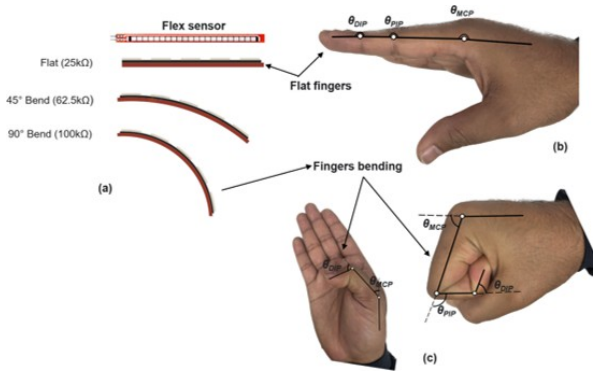

Fig. 3: Estimation of MCP, PIP, and DIP joint flexion/extension using a resistive flex sensor. (A) Schematic representation of the resistive flex sensor, illustrating the transition from the resting position ($0^o$) to full flexion ($90^o$); (B) Depiction of the resting state of MCP, PIP, and DIP joints, where $\theta_{MCP} = 0^o$, $\theta_{DIP} = 0^o$, and $\theta_{PIP} = 0^o$; and (C) Visualization of the fully flexed position of MCP, DIP, and PIP joints.

The flex sensor is integrated with a fixed-value resistor ( 47kΩ) to create a voltage divider. To achieve this, one terminal of the sensor is connected to the power source, while the other connects to a pull-down resistor. Subsequently, the point between the fixed pull-down resistor and the flex sensor is linked to the ADC input of an Arduino. This arrangement generates a variable voltage output that can be effectively monitored by the Arduino's ADC input. It's important to note that the voltage measured pertains to the voltage drop across the pull-down resistor rather than the flex sensor. The voltage divider configuration's output is expressed by the following equation:

$$V_o = V_{cc} \frac{R}{(R + R_{flex})}$$

where $V_0$ is the output voltage, $V_{cc}$ is the system voltage, R is the value of the pull-down resistor, and $R_{flex}$ is the resistance of the flex sensor.

Sensor calibration was performed by obtaining the resistance readings at both 0 degrees and 90 degrees of flexion. The calibration process involved utilizing the map function to translate the sensor's resistance values into corresponding angle measurements based on the data collected at the two reference points. The resistance of the flex sensor changes based on how much it is bent. After a repeated number of trials, the resistance of the sensor may get changed to a normal zero state. To overcome this, the sensors are calibrated.

*Fingertip Force Measurement:* Finger TPS (Total Pressure Sensing) sensors are employed for measuring the forces at the fingertips encountered during grasping objects of DLAs. These sensors incorporate capacitive technology, characterized by the presence of two electrodes that are separated by a compressible dielectric matrix. Notably, capacitive sensors offer distinct advantages, including the fact that the two electrodes remain separated with some distance and function within an elastic range. The selection of sensor is carried out by means of comparison with the available tactile sensors. Table II describes the comparison between the available sensors.

Table II
TACTILE-BASED SENSOR COMPARISON.

| Attribute | Capacitive | Resistive | Piezo-electric |
|---|---|---|---|
| Maximum Range | Good | Good | Good |
| **Sensitivity** | **Excellent** | **Poor** | **Good** |
| Maximum Element Size | Good | Excellent | Poor |
| **Repeatability** | **Excellent** | **Poor** | **Good** |
| Temperature Stability | Excellent | Excellent | Poor |
| Design Flexibility | Excellent | Excellent | Good |

. Capacitive based sensor is selected due to its enhanced sensor's sensitivity, stability, repeatability, and design versatility. Capacitive sensing has thus emerged as the preferred approach for detecting pressure variations arising from contact-based interactions. The characteristics and performance of the capacitive tactile sensor used for fingertip force measurement used in the data glove is shown in Table III.

Table III
SENSOR FOR FINGERTIP FORCE MEASUREMENT.

| Parameter | Value |
|---|---|
| Force Range | 4.4 lbs (2 kg) |
| Sensitivity | 0.1% |
| Signal-to-Noise (SNR) | 1000:1 |
| Gain Non-Repeatability | 1% |
| Linearity | 99% |
| Sensor Thickness | 0.08 in (1.5 m) |
| Weight | 0.12 lbs (55 g) |
| Operating Temperature | 5 - 40 °C |

It works on the basic principle of capacitance, which is defined in terms of equation as,

$$C = \frac{\varepsilon A}{d}$$

where, $C$ is the capacitance, $\varepsilon$ (epsilon) is the permittivity of the material between the plates of the capacitor, $A$ is the area of the capacitor plates and $d$ is the distance between the plates of the capacitor.

When an external force is exerted on the sensor due to the grasp, it results in a decrease in the distance between the two electrodes leading to the increase of the capacitance. These sensors are strategically positioned at the fingertips. The attachment of each sensor to a signal conditioning wrist module is achieved by means of a solitary cable and a 3.5mm connector. The wrist module was linked to a wireless Bluetooth transmitter that was attached to the participant's belt or pocket.

To facilitate the conversion from capacitance to force, which is quantified in Newtons (N), individual calibration of each TPS sensor was carried out. During the calibration procedure, participants were given instructions to gently apply pressure with each finger onto a load cell. The force was incrementally raised until it reached approximately 20N.

sectionExperimental Validation

The sensorized glove developed is to detect finger flexion/extension motion and forces at fingertips during grasping of several household objects from Yale-CMU-Berkeley (YCB) set. An experimental setup is designed for a specific grasping task encompassing the prescribed processes as a benchmark for quantifying performance.

## C. DLA Hand Grasps

Understanding the GTs predominantly employed in daily living activities is vital for evaluating grasp rehabilitation protocols. To this end, the Anthropomorphic Hand Assessment Protocol (AHAP) utilises a total of 25 objects sourced from the publicly accessible YCB Object and Model Set, therefore ensuring the potential for replication in future studies. (see table IV). This set comprises 26 different tasks, encompassing the utilization of eight significant human GTs and two postures that do not involve any grasping. [43]. Grasping capability pertains to the hand's capacity to not only securely grasp a variety of objects used in daily life but also to sustain a stable grip.

Table IV
OBJECTS IN THE EXPERIMENT FROM YCB DATABASE

| Object | Grasp Type | Mass (grams) |
| --- | --- | --- |
| Skillet lid | Hook Grasp (H) | 220.03 |
| Apple | Spherical Grasp (SG) | 158.01 |
| Large Marker | Tripod Pionch (TP) | 12.16 |
| Plate | Extension Grip (EG) | 453.26 |
| Chips Can | Cylindrical Grip (CG) | 107.27 |
| Screwdriver | Diagonal Volar Grip (DVG) | 68.8 |
| Bowl | Lateral Pinch (LP) | 167.67 |
| Small Marker | Pulp Pinch (PP) | 7.67 |
| Switch | Index Pointing (IP) | NA |
| Pitcher base | Hook Grasp (H) | 197.82 |
| Mini Soccer ball | Spherical Grasp (SG) | 16.89 |
| Tuna Can | Tripod Pionch (TP) | 132.59 |
| Craker Box | Extension Grip (EG) | 388.34 |
| Coffee Can | Cylindrical Grip (CG) | 174.42 |
| Spatula | Diagonal Volar Grip (DVG) | 19.72 |
| XS Clamp | Lateral Pinch (LP) | 57.12 |
| Plastic Peer | Pulp Pinch (PP) | 11.01 |
| Plate | Platform | 453.26 |
| Coffee Cup | Hook Grasp (H) | 303.48 |
| Softball | Spherical Grasp (SG) | 59.54 |
| Table Tennis Ball | Tripod Pionch (TP) | 2.74 |
| Tetra Pack | Extension Grip (EG) | 174.94 |
| Power Drill | Cylindrical Grip (CG) | 450.07 |
| Skillet | Diagonal Volar Grip (DVG) | 549.11 |
| Key | Lateral Pinch (LP) | 3.83 |
| Washer | Pulp Pinch (PP) | 2.3 |

26 tasks as shown in Figure 4 is identified. It involves eight different Grasp Types (GTs): pulp pinch (PP), lateral pinch (LP), diagonal volar grip (DVG), cylindrical grip (CG), extension grip (EG), tripod pinch (TP), spherical grip (SG) and hook grip (H). The choices here are grounded in extensive research within the domains of human grasp analysis, prosthetics, and rehabilitation.[12]

The GTs consist of two postures that are not related to grasping: platform (P) and index pointing/pressing (IP). These have been included due to their significance for a multi-grasp prosthetic hand [12]. In order to accommodate the inherent changes in size, shape, weight, texture, and rigidity, a selection of three distinct objects from the YCB collection has been made for each grasp type [44]. This selection is a representative subset of the possible variety of objects encountered in DLAs, considering the available objects within the YCB set's limitations. One objects from the YCB set were chosen for each of the non-grasping postures. In total, 25 objects are utilized for this benchmark, encompassing all the several categories of the YCB set, which include items from food, kitchen, tool, shape, and task.

## D. Subjects

Ten healthy subjects (aged 25-45) are recruited to participate in this study. Subjects are considered to be in good health if they have self-reported the absence of having pain, injury, or disease in their hand, such as arthritis. The experimental procedures adhered to the guidelines outlined in [45], which are designed to prioritize the well-being of the subjects and uphold the credibility of the data, following best practices for research involving human subjects. Regarding the inclusion criteria, all the recruited participants were right-handed and had not reported or observed indications of complications related to cognition or upper extremity function, aligning with the criteria typically applied to individuals with hemiparesis.

## E. Experimental Protocol

Having a standardized set of objects is a beneficial foundation for facilitating consistent and replicable research in manipulation. To ensure uniform and replicable research, it is equally important to establish protocols and benchmarks that provide a clear experimental procedure and reliable quantification methods. As the manipulation research community explores various technical interests, research approaches, and applications, it can be challenging to create task descriptions that can accommodate the diverse interests and remain relevant over time due to the constantly evolving nature of the field. With the aim of promoting diverse types of manipulation, the experiment was designed and performed considering various factors such as *shape*, *size*, *weight*, *grasp gesture*, and *stiffness*. Data was collected for each subject involving a repeated number of trials of grasp with the various GTs. Different daily use objects were chosen for experiment. All selected objects for the experiment with grasp postures are shown in figure 4.

The camera is set at a distance of 60 cm from the object for capturing the grasping trial, each was of 30 seconds. The object is put on a table in front of the camera. The subjects were instructed to keep their gloved hand in an initial resting position, with the palmer side facing downwards on the table. Subsequently, they were directed to grasp and elevate the target object to a height of 15 cm above the table and maintain a stable "hold" for 20 seconds. The scan rate of both sensors is set to 40 Hz. The mass of the household objects selected is taken below 1.5 kg as the force sensor is calibrated to 15 newtons.

The movement of each trial of DLA was divided into four phases, defined by the marked events as follows:

- **Approaching**, until touching the object to be grasped.
- Touching the object for **Grasping**.
- **Lifting** the object for 10 cm height.
- **Holding** the object till the grasp force at each finger tips becomes constant.

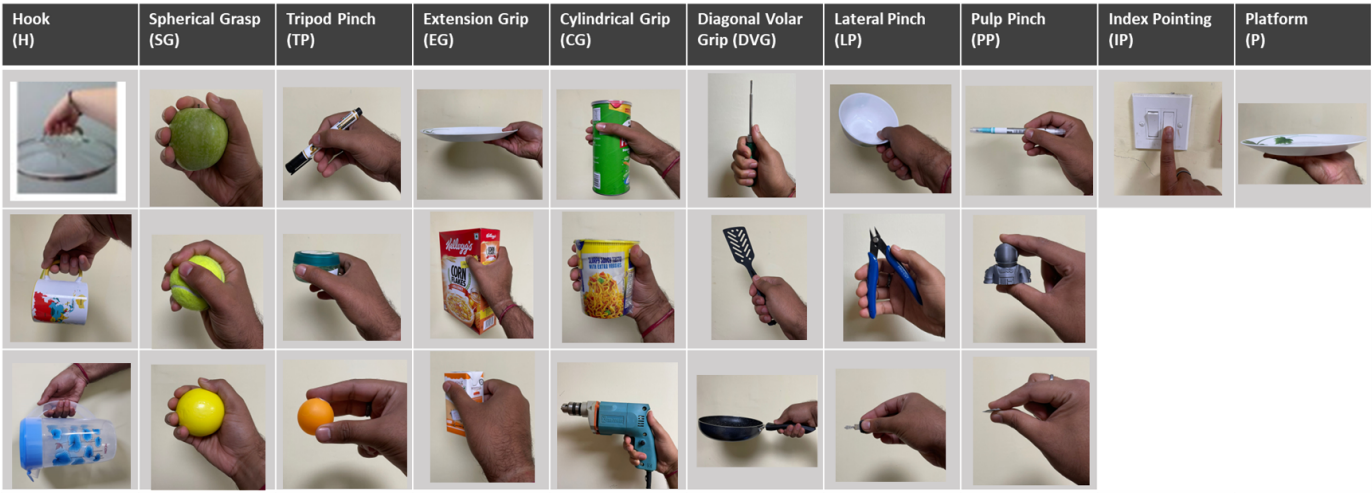

Fig. 4: Grasp types and objects (YCB set) used in the Experimental protocol.

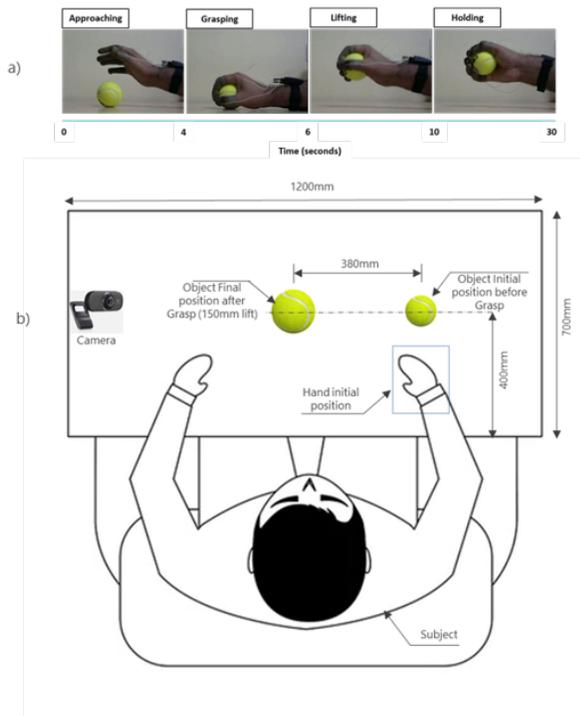

Fig. 5: a) Experiment Timing Diagram b) Experimental Setup

The above-mentioned process of the experiment and the experimental setup is described in figure 5.

During the experiment, it was observed that on certain trials, the fingers establish contact with the object, but failed to elicit discernible sensor readings. It is due to the fact that grasp was undertaken by the finger pads and the palm without the involvement of the distal phalanges and the fingertips, wherein the force sensors are located. In order to avoid such scenarios experiment was done ensuring that the grasping is completed involving the distal phalanges of the fingers. As the participants are instructed to execute their preferred natural grasps without imposing any additional constraints, a diverse range of grasp postures is recorded for the same object.

*F. Human Grasp Recordings- Data Collection*

The data collection is done with different subjects with all the 25 objects with 10 trials. The data set in .csv format consists of the value of grasp force (in newtons) and finger bending angles (in degrees) at the Thumb, Index, Middle, Ring and Pinky finger in each trial. The tactile and bending angle data contains 5 channels and 1200 steps within 30 seconds. The video data is synchronized with the tactile data during each trial. A sample data for the synchronised values of grasp forces and postures while grasping a spherical object is shown in the figure 6.

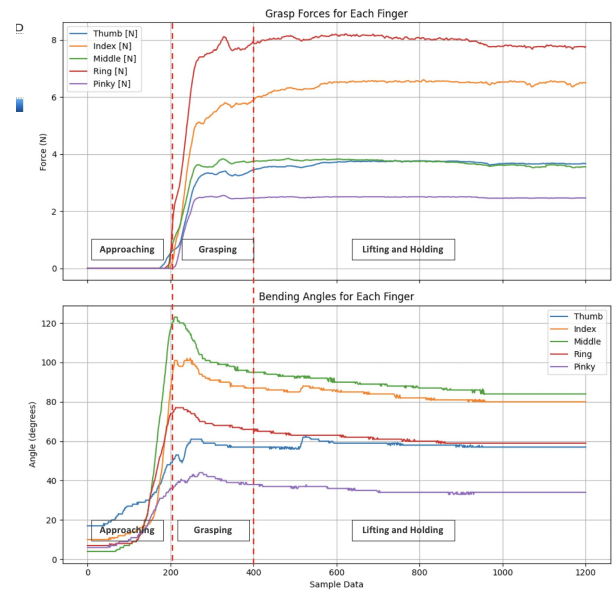

Fig. 6: Sample Data of Grasp Forces and Finger Bending Angles while grasping Spherical Object.

## V. RESULTS AND DISCUSSION

### A. Finger Bending Angle and Finger-tip Force Relationship

Comprehending human grasping serves as a foundational step in shaping the grasping capabilities of robotic assistive devices. Human grasping possesses remarkable adaptability and offers substantial benefits in terms of angle precision, force control, and slip prevention. A more profound understanding of grasping, particularly the interplay between finger bending angles and grasp force magnitude, can provide valuable insights for enhancing robot grasping, setting thresholds, and analyzing outcomes. Figure 6 shows the relationship between the force and the bending angle of the five fingers during a grasping process.

Figure 6 shows that the grasping process is divided into mainly three phases, the approaching phase, the grasping phase and then the lifting and holding phase. In the approaching phase, once the fingers start bending to approach the object for grasping, the angle values start increasing in synergy but the force values remain zero. Once the fingers touch the object and grab it, the force values start increasing from zero value and at this stage, the angle values start stabilising its value. Once the grasping phase gets over, the force values also become constant. In the lifting phase, the angle and force values slightly adjust their value and at last in the holding phase, all the values get almost stabilized. The same pattern of force angle relationship is followed by all the fingers while grasping any object.

### B. Grasp posture and force representation using Radar Charts

Figure 7 shows the radar plots showcasing the grasp posture representation for different GTs based on empirical relations between the MCP, PIP, and DIP joints as discussed in section III-A. Each chart depicts the joint angles of the fingers in a specific GTs, with each joint angle corresponding to a spoke on the radar chart. The length of each spoke represents the magnitude of the joint angle, providing a visual depiction of the hand posture for each grasp. These spider radar charts offer valuable insights into the kinematic patterns and coordination of the hand during various GTs. From the radar charts, it has been observed that in most of the GTs the maximum flexion is at the MCP joints of each finger and the DIP joints contributes the least flexion in all the type of grasps. Also, the index and middle fingers have shown an overall maximum flexion and in contrast, the ring and pinky fingers have their least contribution of flexion while grasping. This indicates the importance of the index and middle fingers as they play a very crucial role in grasping. Meanwhile, the ring and pinky finger flexions indicate their role as supporting fingers while grasping for proper balancing and stability of an object. The thumb has moderate flexion in each grasp type as it has only two joints which restrict its flexion, moreover, it acts as an opposing finger from the other side of the four fingers while grasping an object. Similarly, the radar plot for the mean of grasp forces at the static phase of grasp (holding) of the grasp for grasp type is shown in figure 8. It shows that, in almost each of the GTs the thumb finger contributes a larger force than the other fingers as it has to compensate the opposition forces applied by them. The index and middle finger contribute moderate force and it mainly helps in maintaining grip strength. The ring and pinky fingers contribute the least amount of force and the magnitude of their forces are statically equivalent, as these fingers mainly contribute to maintaining the stability of the grasp. Also, as per [8] and [9] classification of grasp types in power and precision grasp, the grasp types SG, CG and EG under the power grasp types have greater area enclosed in the pentagonal chart than the precision grasps like TP, LP, PP and DVG. The finding of this study have the potential to be applied in the design and development of grasp controllers for five-fingered robotic and or mimicking human grasping behaviour.

### C. Influence of object mass on Grasp Force

To successfully lift the object from the table, the grasp force applied by the fingers must compensate for the object's weight. The magnitude of the overall grasp force is contingent upon the weight of the object. This is clearly confirmed by our study on the grasp forces.

The present study investigates the grasp force characteristics across different GTs as a function of the mass of the object. The study employed linear interpolation techniques averaged over all measured grasp forces based on the object weight for a particular grasp type to estimate the grasp forces at different mass levels, allowing for a detailed examination of the force-mass relationship for each grasp type. The resulting plots clearly illustrated the force variations and trends exhibited by each finger for different GTs. Figure 9, demonstrates the force profiles for each finger across multiple GTs. We also investigated the total force exerted during different GTs as a function of the mass of the object.

The magnitude of the grasp force varies depending on the weight of the object. Heavier objects require a more substantial grasp force, leading to significant variations in fingertip forces among subjects based on the dynamic properties of the chosen grasp type. Among the grasp types, Spherical Grasp (SG) demonstrated the highest maximum force, reaching a peak of 9.96 N at a mass of 158 g, while Lateral Pinch (LP) exhibited the lowest minimum force, 1.97 N at a mass of 167 g. These findings shed light on the relationship between grasp force and object mass.

### D. Fingers Correlation Coefficient Analysis

To gain deeper insights into the dynamics of human hand grasping, we examine the correlations between grasp forces and finger bending angles. Our analysis commences with the evaluation of the bending angles for each finger, followed by an investigation into the fingertip forces throughout the grasping process. Subsequently, we compile the dataset encompassing angle and corresponding force data for each finger across various GTs, each involving three distinct objects.

To extract synergies, it's essential to identify similarities in force behaviour and explore correlations between contact forces distributed across the fingertips and bending angles along the dorsal side of the hand while grasping. In our comprehensive analysis, we've uncovered substantial correlations between fingers by examining forces and angles across diverse objects and trials. These correlations are evaluated using the Pearson

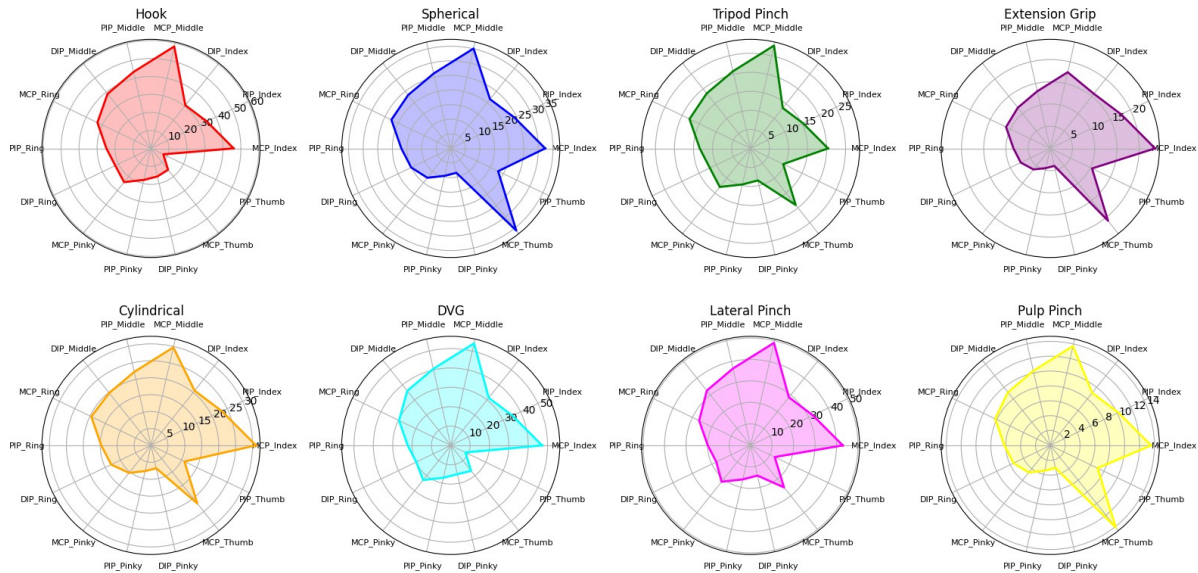

Fig. 7: Radar plots illustrating the joint angles of the metacarpophalangeal (MCP), proximal interphalangeal (PIP), and distal interphalangeal (DIP) joints for all five fingers of the human hand for different grasp types

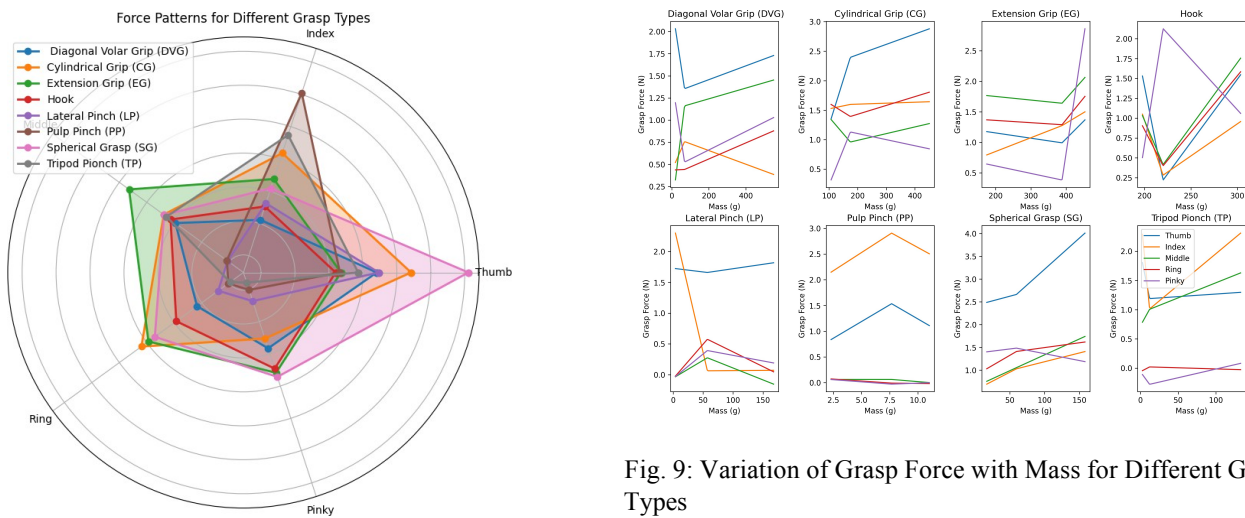

Fig. 9: Variation of Grasp Force with Mass for Different Grasp Types

Fig. 8: Radar plot illustrating the grasp forces for all five fingertips of the human hand for different grasp types

Correlation Coefficient, enabling us to assess the relationship between two contact forces or finger-bending angles.

A positive correlation means that as one variable increases, the other variable tends to increase as well. A negative correlation means that as one variable increases, the other variable tends to decrease. The closer the correlation coefficient is to 0, the weaker the relationship between the two variables. The full correlation matrix is shown in figure 10.

In the context of grasp forces, this could mean that as the grasp force at one fingertip increases, the grasp force at another fingertip decreases. This could happen, for example, if the object being grasped is asymmetrical or if the grasp posture is not optimal, leading to imbalanced grasp forces. However, it is worth noting that a negative correlation between grasp forces is not necessarily the norm and may not always be desirable in functional grasping tasks.

The correlation coefficients between grasp force and grasp posture for each pair of fingers are analyzed for understanding the relationship between these variables (see figure 10). The maximum and minimum correlation values of grasp postures and forces mentioning the particular grasp type are shown in Figure 11. The results reveal that, for the pair of Index and Middle fingers, a strong positive correlation was observed between grasp force and grasp posture, with a maximum value of 0.95. This indicates that as the grasp force increases, there is a corresponding change in the grasping posture, suggesting a coordinated behaviour between these fingers. The grasp types associated with the maximum correlation force and posture were Spherical Grasp (SG) and Extension Grip (EG), respectively. In contrast, the Index and Pinky fingers pair showed a moderate positive correlation between grasp posture and grasp force, reaching the maximum correlation coefficient of 0.740. This observation implies a less pronounced relationship between grasp posture and force compared to the pair of Index-Middle fingers. The grasp types that exhibited the maximum correlation force and posture were identified as Spherical Grasp (SG) and Diagonal Volar Grip (DVG), respectively.

Similarly, the pair of Index-Ring fingers showed a strong positive correlation, with a maximum correlation coefficient of 0.898. This suggests a significant synergy between grasp force and grasp posture when considering the Index-Ring pair of fingers together. The Middle-Pinky pair of fingers showed a considerable positive correlation, with a maximum correlation coefficient of 0.686. This indicates a relatively weaker but still noticeable relationship between grasp force and grasp posture for these fingers. The Middle-Ring pair of fingers showed a very strong positive correlation, with a maximum value of 0.969. This indicates a highly coordinated relationship between grasp force and grasp posture when considering the Middle and Ring fingers together. For the Thumb-Index, Thumb-Middle, Thumb-Pinky, and Thumb-Ring finger pairs, varying degrees of positive correlations were identified. The maximum correlation coefficients ranged from 0.868 to 0.889, indicating significant coordination between grasp force and grasp posture for these pairs of fingers. Overall, the analysis of correlation coefficients highlights the cooperative behaviour and synergy between the pair of fingers during a grasping process. The study provides significant insights into the interdependency between grasp force and grasp posture, emphasizing the coordinated nature of hand movements during object grasping and manipulation.

### E. Grasping Analysis - Grasp Synergy

The ability of humans to grasp objects is intricately linked to the concept of synergy, particularly within the domain of movement control and coordination. Synergy includes how the muscles, joints, and sensory feedback in the hand coordinate to make a functional and efficient grasp. While grasping, the human hand must dynamically adjust its posture, finger positions, and forces applied to the object being grasped. The concept of synergy serves to elucidate how the numerous DoFs in the hand are coordinated and streamlined into meaningful movement patterns for a successful grasp.

Dimensionality reduction techniques are important in understanding grasp synergy, particularly in encoding human grasp postures and analyzing the forces involved. Principal Component Analysis (PCA) is a widely used method that extracts the most significant features or components from the dataset and represents it in a lower-dimensional space.

In order to comprehend the grasp postural and force synergy, the grasp dataset was visualized. PCA was utilised to analyze the grasp forces and hand configurations separately. PCA enables a reduction in dimensionality, providing insights into the kinematic and dynamic aspects of the hand while grasping. The goal is to identify the key factors or principal components (PCs) that consist of the majority of the variance within the dataset. By doing so, PCA enables a more concise representation of the grasp data while preserving the essential information necessary for understanding grasp synergy.

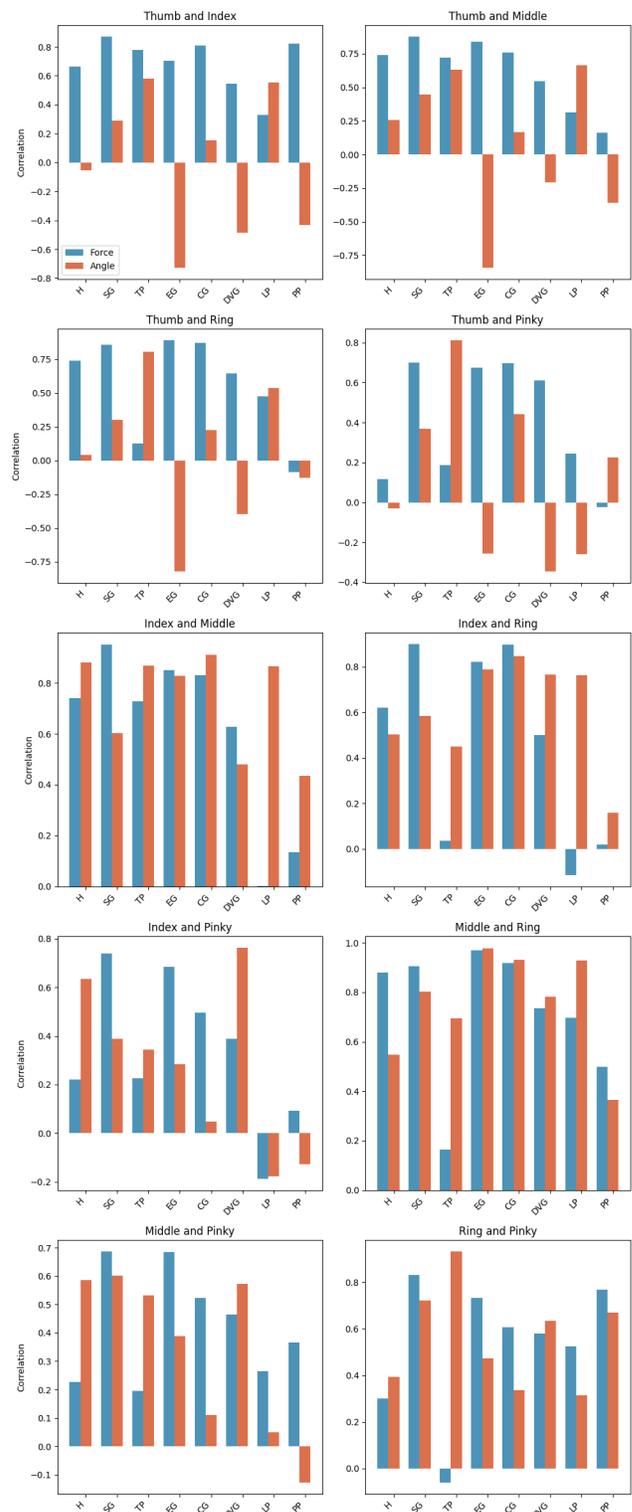

Fig. 10: Correlation coefficient for Grasp forces and postures between the pairs of fingers for different grasp types

The results of the Principal Component Analysis (PCA) for grasp posture reveal important insights into the dimensionality of the dataset. The cumulative variance explained by the principal components provides a measure of how much information is retained as we increase the number of components. In this analysis, the optimal number of components for the dataset is determined through the "Elbow Method". It is a graphical technique to determine the optimal number of PCs in PCA. It analyses the variance explained by each of the PCs and identifies the point at which adding more PCs does not significantly increase the explained variance. It starts by performing the PCA to obtain eigenvalues and their corresponding explained variances. This represents the amount of data variance explained by each PC. Subsequently, a scree plot is generated, where the x-axis denotes the number of PCs, and the y-axis shows the cumulative explained variance as shown in figures 12 and 13. The "elbow" point on this plot corresponds to the point where adding more PCs yields declining returns in terms of explained variance. This "elbow" represents the optimal number of components as highlighted by the dotted red lines. The benefit of using this method lies in its simplicity in its representation, quick visualization for exploratory data analysis, and intuitive approach to dimensionality reduction, rendering it a valuable tool for exploratory data analysis.

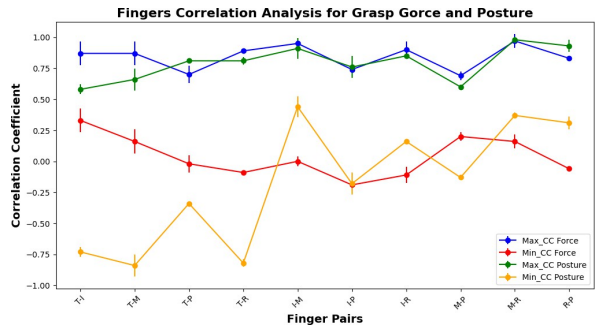

Fig. 11: Maximum-Minimum Correlation coefficient of Grasp Force and Posture

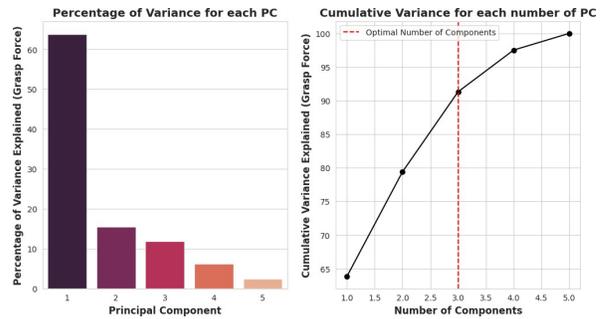

Fig. 13: Explained and Cumulative variance ratios of Grasp Forces

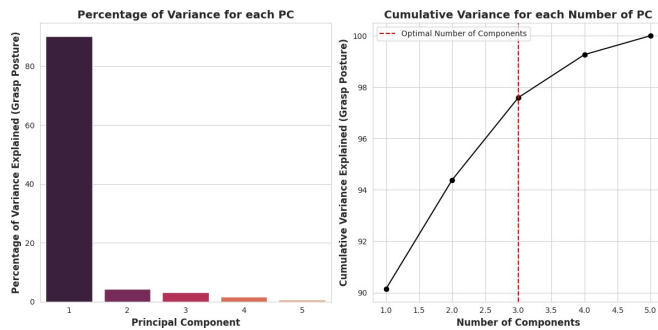

Fig. 12: Explained and Cumulative variance ratios of Grasp Posture

The optimal number of PCs for the grasp posture is determined as 3 (refer Figure 12). When considering only one principal component, approximately 90.14% of the variance in the data. Adding a second component increases the cumulative variance to 94.39%, indicating that the second component captures additional important information. Finally, with the inclusion of the third component, the cumulative variance reaches 97.59%, suggesting that the third component contributes further to understanding the variation in grasp postures.

While the kinematic hand configuration can be effectively represented using PCA, the grasp forces pose a greater challenge due to their higher diversity and complexity. The results of the PCA for grasp forces also indicate that the optimal number of components for the grasp dataset is 3 (refer Figure 13), based on the cumulative variance explained. When considering only one component, it accounts for 63.87% of the total variance in the grasp force dataset, implying that a single component captures a significant portion of the underlying variability in the force exerted during grasping. Expanding to 2 components increases the cumulative variance to 79.42%. The inclusion of the second component enhances the amount of variance and provides a more comprehensive representation of the grasp force patterns. However, the cumulative variance reaches 91.31% when considered with 3 components. This indicates that there is a substantial contribution of the third component by giving additional information on the dataset, enhancing the understanding of the grasp force dynamics. By considering 3 components, the grasp force patterns can be effectively represented with a high level of accuracy and capture the variability in the data significantly. Hence, the PCA results highlight the importance of considering multiple components as PC to fully capture the complexity and nuances of grasp force analysis, enabling a more holistic understanding of the forces involved in the grasping process. By utilising dimensionality reduction techniques like PCA, researchers can gain a deeper understanding of grasp synergy in terms of the kinematic and dynamic aspects. This methodology facilitates the identification of underlying patterns and contributes to the advancement of five-fingered robotic systems capable of mimicking human grasping capabilities.

In this study, t-SNE (t-Distributed Stochastic Neighbor Embedding) [46], was employed to visualize and cluster the grasp force and posture data. t-SNE is widely employed for reducing high-dimensional data resulting in a representation of the data in two or three dimensions while preserving the inherent structure of the dataset. One of the key benefits of t-SNE is its ability to preserve the local and global similarities between data points, enhancing its effectiveness in revealing underlying patterns and clusters in complex datasets. By projecting the high-dimensional grasp force and posture data onto a lower-dimensional space, t-SNE creates a scatter plot which enables intuitive visualization and interpretation. The utilisation of t-SNE facilitates to identification of distinct groups or clusters of similar grasp force and posture patterns. This visualization aids in the understanding of inherent variability and interconnections between different grasp types. Moreover, t-SNE preserves the topological structure of the data, ensuring that nearby points in the original high-dimensional space retain their proximity in the reduced-dimensional space. This study using t-SNE provides a comprehensive visualization of the grasp force and posture data, allowing for a deeper understanding of the underlying structures and relationships between the different grasp types. The use of t-SNE facilitates the identification of clusters, which enhances the analysis and interpretation of the grasp force and posture data.

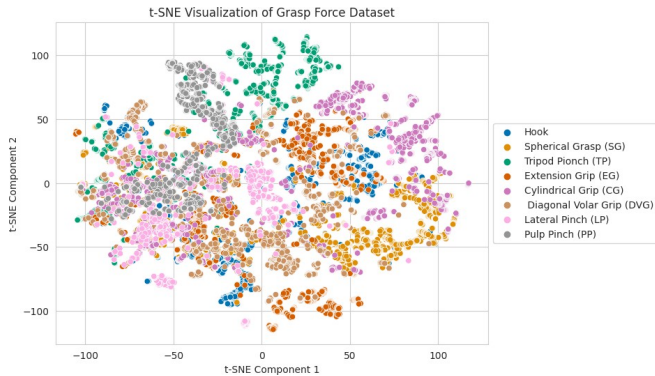

Fig. 14: t-SNE Visualization of Grasp Forces

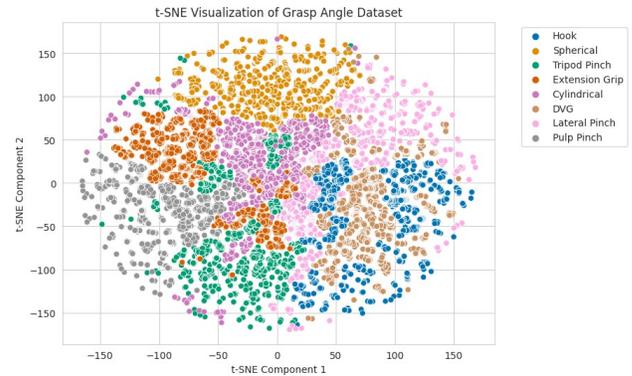

Fig. 15: t-SNE Visualization of Grasp Postures

The t-SNE clustering representation of the grasp dataset provides insights into the relationships between different GTs. Through the scatter plot, we can observe how the grasp postures and grasp forces cluster together, revealing patterns and similarities among the different GTs. In the t-SNE scatter plot Fig. 15 and Fig. 14, it can be observed that grasp types like the Hook Grasp, Spherical Grasp, Cylindrical Grasp, Diagonal Volar Grip, and Lateral Pinch tend to cluster together. This clustering pattern indicates that these grasp types share similarities in terms of their posture and force characteristics. These grasp types typically involve a grip with a larger force (power grasp), and the t-SNE clustering representation reflects this by grouping them together based on their similarities in the postural and force attributes. On the other hand, grasp types like the Tripod Grasp, Extension Grip, and Pulp Pinch are clustered separately from the power grasp types. This observation of the clustering pattern indicates that these precision grasp types exhibit unique posture and force characteristics which distinguish them from the power grasp types. These precision grasp types often involve more delicate movements and precise force control of the fingers, and the t-SNE clustering representation reflects this by grouping them together based on their shared attributes of grasp posture and force. Overall, the t-SNE clustering representation of the grasp dataset facilitates the visual understanding of the relationships between grasp postures and grasp forces among the different GTs. The identified clusters of grasp types exhibit similar grasp characteristics, which enables a better understanding of the underlying patterns and synergies in human grasp.

## VI. CONCLUSION

The design and development of an instrumented data glove is reported in this paper. The glove is developed using 3D printing technology with a flexible material *i.e.* Thermoplastic Polyurethane. The data glove measures the kinematics and dynamics of the human hand in terms of grasp postures and forces. A comprehensive study involving a variety of objects used in daily living activities consisting of eight different grasp types with three objects for each type involving ten healthy subjects was undertaken. The dataset is analysed for postural and force synergy. The research focused on the examination of the grasp force and bending angle relationship, the influence of mass on the grasp force, finger cooperation characteristics through correlation coefficients for a pair of fingers in terms of finger bending movements and the force applied by each fingertip while grasping. Analyzing the relationship between finger bending angles and grasp forces could contribute to robotic grasping, particularly setting thresholds, and exploiting results for design and control. The grasp force and posture correlation analysis reveals important insights into their relationship in human hand movements during object manipulation. The findings highlight the cooperative behaviour and synergy between fingers during grasping. Synergy plays a crucial role in movement control and coordination. It involves the coordinated actions of muscles, joints, and sensory feedback to achieve effective grasping.

The current study on grasp synergy is of relevance in the design and control of prosthetic hands and exoskeletons for rehabilitation. By understanding the force and bending angle relationship, together with the coordination between fingers, one can optimize the exoskeleton's design to closely mimic the natural hand movements and grasp forces. The findings from the study can guide the selection and placement of sensors and actuators within the assistive devices to accurately capture and replicate the desired grasp patterns. Additionally, the grasp synergy analysis using PCA and the t-SNE scatter plot can aid in developing control algorithms that enable the devices to adapt to different grasp tasks and provide appropriate force assistance based on the user's intentions. This is part of on-going research.